\documentclass[runningheads]{llncs}
\usepackage{graphicx}
\usepackage{hyperref}
\usepackage{amssymb}
\usepackage{amsmath}
\usepackage{relsize}
\usepackage{booktabs}
\usepackage[dvipsnames]{xcolor}
\usepackage{chemformula}

\makeatletter
\AtBeginDocument{%
  \def\doi#1{\url{https://doi.org/#1}}}
\makeatother

\newcommand{\etal}{\textit{et al. }}
\renewcommand\#{\protect\scalebox{0.8}{\protect\raisebox{0.4ex}{\char"0023}}}

\newtheorem{assumption}{Assumption}
\newcommand{\megan}{\textsc{Megan}}
\newcommand{\meganII}{\textsc{Megan2}}

\newcommand{\gatII}{\textsc{Gat}v2}

\DeclareMathOperator*{\argmin}{arg\,min}

\newcommand{\rsquare}{R^{2}}

\newcommand{\circla}{\textcircled{\raisebox{-.1pt} {\smaller a}}}
\newcommand{\circlb}{\textcircled{\raisebox{-.9pt} {\smaller b}}}
\newcommand{\circlc}{\textcircled{\raisebox{-.3pt} {\smaller c}}}

\newcommand{\inputElem}{\mathcal{X}}
\newcommand{\fragmentElem}{\mathcal{\hat{X}}}
\newcommand{\channelNum}{K}
\newcommand{\channelIndex}{k}
\newcommand{\targetElem}{\mathcal{Y}}
\newcommand{\targetPred}{\mathbf{y}^{\mathrm{pred}}}

\newcommand{\targetNum}{C}
\newcommand{\params}{{\pmb{\theta}}}
\newcommand{\layerNum}{L}
\newcommand{\layerIndex}{l}
\newcommand{\batchNum}{B}
\newcommand{\batchIndex}{b}

\newcommand{\embedding}[1]{\mathbf{h}^{(#1)}}
\newcommand{\graphEmbedding}{\mathbf{h}}
\newcommand{\embeddingDim}{D}
\newcommand{\projection}[1]{\mathbf{z}^{(#1)}}
\newcommand{\graphProjection}{\mathbf{z}}
\newcommand{\projectionDim}{P}

\newcommand{\graph}{\mathcal{G}}
\newcommand{\subgraph}{\mathcal{\hat{G}}}
\newcommand{\nodes}{\mathcal{V}}
\newcommand{\nodeNum}{V}
\newcommand{\nodeAttr}[1]{\mathbf{H}^{(#1)}}
\newcommand{\nodeImportances}{\mathbf{V}^{\mathrm{im}}}
\newcommand{\nodeMask}{\mathbf{\hat{V}}}
\newcommand{\edges}{\mathcal{E}}
\newcommand{\edgeNum}{E}
\newcommand{\edgeAttr}[1]{\mathbf{U}^{(#1)}}
\newcommand{\edgeImportances}{\mathbf{E}^{\mathrm{im}}}
\newcommand{\edgeAttention}[1]{\mathbf{A}^{(#1)}}
\newcommand{\edgeMask}{\mathbf{\hat{E}}}

\newcommand{\lossPred}{\mathcal{L}^{\mathrm{pred}}}
\newcommand{\lossExpl}{\mathcal{L}^{\mathrm{expl}}}
\newcommand{\lossSpar}{\mathcal{L}^{\mathrm{spar}}}
\newcommand{\lossContr}{\mathcal{L}^{\mathrm{contr}}}

\newcommand{\concept}{\mathcal{C}}

\begin{document}

\title{Global Concept Explanations for Graphs by Contrastive Learning}
\titlerunning{Contrastive Global Graph Concept Explanations}
%
\author{ %
Jonas Teufel\inst{1}\orcidID{0000-0002-9228-9395} %
\and %
Pascal Friederich\inst{1}$^{,}$\inst{2}$^{\dagger}$\orcidID{0000-0003-4465-1465}
}
\authorrunning{Teufel et al.}
%
\institute{ %
Institute of Theoretical Informatics, Karlsruhe Insitute of Technology,\\
Kaiserstr. 12, 76131 Karlsruhe, Germany\\
\email{jonas.teufel@kit.edu}
\and
Institute of Nanotechnology, Karlsruhe Insitute of Technology,\\
Kaiserstr. 12, 76131 Karlsruhe, Germany\\
\email{pascal.friederich@kit.edu}
}

\maketitle              
%
\begin{abstract}
Beyond improving trust and validating model fairness, xAI practices also have the potential to recover valuable scientific insights in application domains where little to no prior human intuition exists. To that end, we propose a method to extract global concept explanations from the predictions of graph neural networks to develop a deeper understanding of the task's underlying structure-property relationships. We identify concept explanations as dense clusters in the self-explaining \megan~model's subgraph latent space. For each concept, we optimize a representative prototype graph and optionally use GPT-4 to provide hypotheses about why each structure has a certain effect on the prediction.
We conduct computational experiments on synthetic and real-world graph property prediction tasks. For the synthetic tasks we find that our method correctly reproduces the structural rules by which they were created. For real-world molecular property regression and classification tasks, we find that our method rediscovers established rules of thumb. More specifically, our results for molecular mutagenicity prediction indicate more fine-grained resolution of structural details than existing explainability methods, consistent with previous results from chemistry literature. Overall, our results show promising capability to extract the underlying structure-property relationships for complex graph property prediction tasks.
\end{abstract}
\vspace*{-10pt}

\renewcommand{\thefootnote}{$\dagger$}
\footnotetext{Corresponding author.}
\renewcommand{\thefootnote}{\arabic{footnote}}

%

\section{Introduction}

The practice of explainable artificial intelligence (xAI) is often expected to improve trust during human-AI interactions, provide tools for model debugging, validate model fairness, and comply with anti-discrimination laws \cite{doshi-velezRigorousScienceInterpretable2017}. These merits are of high interest in application domains where humans already have solid intuitions and expectations about how a model \textit{should} behave, such as image and natural language processing tasks.\\
We argue that in applications in which little to no prior intuition about a task exists, xAI practices can be used to extract knowledge from a high-performing AI model and to gain new insights about the underlying tasks. Especially important application domains in which this is the case are chemistry and material science. In recent years, graph neural networks (GNNs) have proven to be a valuable tool for producing accurate property predictions from molecular graph representations \cite{reiserGraphNeuralNetworks2022a}. Despite access to accurate predictive models, the underlying working mechanisms of many of these tasks remain open questions.\\
In most application domains, including graph processing, there already exist various xAI methods that help to make AI predictions more transparent. However, most of these methods are focused on local explainability, which aims to supply additional information and explanations for each individual prediction. As previous authors have stated already \cite{yuanExplainabilityGraphNeural2022,kakkadSurveyExplainabilityGraph2023a}, there is a lot of additional potential in generating global explanations, rather than local instance-level explanations. In contrast to purely local explanations, global explanations aim to provide additional information about the behavior of the model as a whole. 
Here, we present a method to generate global concept explanations for graph property prediction tasks. The aim of these global explanations is to approximately extract the general structure-property relationships that govern the model's decisions. For this purpose, we propose \meganII---an extension of the recently proposed multi-explanation graph attention network (\megan) \cite{teufelMEGANMultiexplanationGraph2023a} architecture. Our modifications include an additional contrastive representation learning objective which promotes the model's latent space of subgraph embeddings to be aligned according to the structural similarity of the corresponding motifs. Explanations are then generated by applying clustering methods to this latent space.
The extracted clusters consisting of similar subgraph motifs can be interpreted as overarching \textit{concepts} discovered and exploited by the model. By analyzing the prediction contributions of the individual cluster members, it is also possible to determine each concept's average influence on the prediction outcome. Furthermore, we use a genetic algorithm optimization to determine a small and representative prototype graph for each cluster. Additionally, a suitable representation of this prototype graph can then be given to a large language model such as GPT-4 \cite{IntroducingChatGPT,bubeckSparksArtificialGeneral2023} to provide an initial hypothesis about possible causal relationships behind the identified structure-property correlation.\\
We experimentally validate the proposed framework for global explainability on several synthetic graph prediction tasks and real-world molecular property prediction tasks. On synthetic datasets for graph regression and classification, we show that our method can correctly reconstruct the structure-property relationships based on which the datasets were created. Furthermore, we validate our method on two molecular property prediction datasets for the prediction of water solubility and the classification of mutagenicity. For these real-world datasets, our method re-discovers known rules of thumb about the underlying molecular properties. Specifically for the mutagenicity prediction we find that our method produces significantly more fine-grained explanations than previously published methods for global graph explainability which are consistent with previously published work from the chemistry literature. This opens wide applicability to many tasks in graph classification and regression, specifically in chemistry and materials science, to potentially derive unknown structure-property relationships.\\

\textbf{Contribution. } Compared to the previously published literature on the topic of global concept explainability for graph neural networks our main contributions are the following:
\begin{itemize}
\item[$\bullet$] Previous work focuses entirely on graph classification tasks. However, regression tasks are equally important---especially considering the various material property prediction applications in chemistry and material science. We show that our method can produce reasonable explanations for classification and regression tasks.
\item[$\bullet$] Previous work either uses partitioning methods \cite{magisterGCExplainerHumanintheLoopConceptbased2021} or a concept-bottleneck architecture \cite{azzolinGlobalExplainabilityGNNs2022b}, which requires the number of concepts to be configured beforehand. We argue that this presents a limitation, since a good estimation of the number of concepts may not be possible for unknown tasks. By building on a density-based clustering method, our method can more naturally discover the appropriate number of explanations for each task dynamically.
\item[$\bullet$] We present an entirely self-explaining method, where the local, as well as global explanations, are derived directly from a single attention-based model. A self-explaining approach aims to alleviate the recently brought-up conceptual problem of trying to "explain a black box with another black box" which increasingly complex post-hoc approaches are facing \cite{rudinStopExplainingBlack2019,bordtPostHocExplanationsFail2022}.
\end{itemize}

\begin{figure}[t!]
    \centering
    \includegraphics[width=\textwidth]{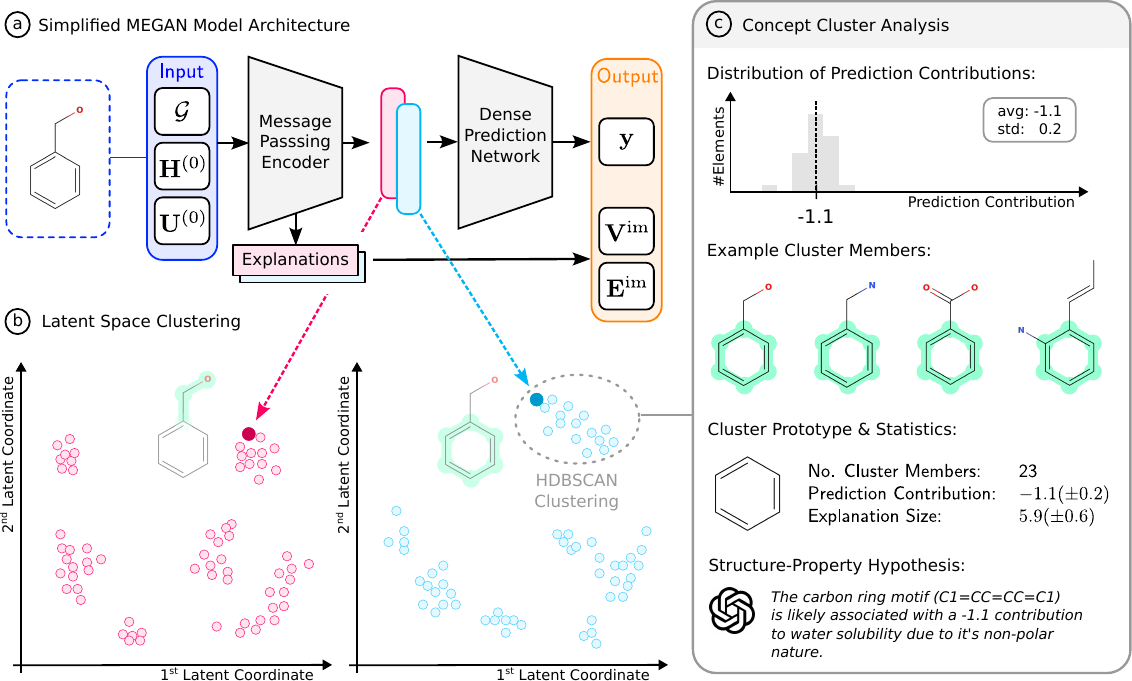}
    \caption{Visual overview of the proposed global explanation method. \circla~Simplified MEGAN model architecture. The message-passing encoder creates explanation masks along multiple channels (pink/blue), which then result in individual subgraph latent representations for each channel. \circlb~In each channel's subgraph latent space, a clustering algorithm is used to find elements in the dataset that exhibit structurally similar explanation masks. \circlc~Each cluster is analyzed regarding its members and results are presented to the user in the form of an automatically generated report.}
    \label{fig:basic-overview}
    \vspace*{-0.2cm}
\end{figure}

\section{Related Work}

\textbf{Graph Explainability. } Graph neural networks have proven to be useful tools to handle predictive tasks on graph-structured data, such as molecular and material property prediction \cite{reiserGraphNeuralNetworks2022a}.
In parallel, xAI methods are developed to make the predictions of these complex models more transparent and interpretable to their human users. Yuan \etal \cite{yuanExplainabilityGraphNeural2022}, and more recently Kakkad \etal \cite{kakkadSurveyExplainabilityGraph2023a}, provide a comprehensive overview of the existing literature on explainability methods specifically for graph neural networks. The majority of this literature is focused on local attributional explanation methods, that aim to provide additional information about the reasoning for each prediction. Many existing methods for local explainers, such as GraphLIME \cite{huangGraphLIMELocalInterpretable2020}, GNN GradCAM \cite{popeExplainabilityMethodsGraph2019} and GraphShap \cite{perottiGRAPHSHAPExplainingIdentityAware2023}, are adaptations of previously existing work from the image and text processing domains. Other notable local graph explanation methods include GNNExplainer \cite{yingGNNExplainerGeneratingExplanations2019b}, PGExplainer \cite{luoParameterizedExplainerGraph2020}, SubgraphX \cite{yuanExplainabilityGraphNeural2021b}, and more recently the self-explaining \megan~model architecture \cite{teufelMEGANMultiexplanationGraph2023a}. Besides attributional explanation masks, counterfactual explanations are another popular modality for local explanations of graph neural networks \cite{lucicCFGNNExplainerCounterfactualExplanations2021,tanLearningEvaluatingGraph2022}.
Contrary to local explainers, methods of global explainability aim to provide information about large sections of the model's decision space at once---largely decoupled from individual predictions. Although global explainability is less intensively explored, existing work can roughly be split into generative explanations \cite{yuanXGNNModelLevelExplanations2020} prototype-based \cite{daiPrototypeBasedSelfExplainableGraph2022,zhangProtGNNSelfExplainingGraph2022} and concept-based explanations \cite{magisterGCExplainerHumanintheLoopConceptbased2021,azzolinGlobalExplainabilityGNNs2022b}.\\
In parallel to the development of the core explainability mechanisms, another area of ongoing research is that of AI visual analytics systems \cite{hohmanVisualAnalyticsDeep2019}. In this line of work, GNNLens \cite{jinGNNLensVisualAnalytics2023} and CorGIE \cite{liuVisualizingGraphNeural2021} are recently proposed visual analytics systems specifically created for the exploration of GNN predictions.\\ 

\textbf{Concept Explanations. } An early notion of concept explanations is presented by Kim \etal \cite{kimInterpretabilityFeatureAttribution2017} who developed their framework for testing with concept activation vectors (TCAV) as an alternative to saliency maps for visual explanations. However, for this initial formulation, the concepts have to be manually defined to test their relevance to the prediction outcome. Ghorbani \etal \cite{ghorbaniAutomaticConceptbasedExplanations2019} extend this method with their automated concept explanation (ACE) algorithm. The authors automatically assemble concept clusters by clustering image fragments according to their similarity in the model's latent space. These methods are subsequently extended and refined in various image-processing \cite{felCRAFTConceptRecursive2023,zhangInvertibleConceptbasedExplanations2021} and text-processing applications \cite{shiCorpuslevelConceptbasedExplanations2022,jourdanCOCKATIELCOntinuousConcept2023}.\\

\textbf{Graph Concept Explanations. } Even though concept-based explanations have mainly been applied to the image- and text-processing domains, there already exists some previously published work on concept-based explainability for graph neural networks as well. To our knowledge, Magister \etal \cite{magisterEncodingConceptsGraph2022a} first introduced concept explanations to the graph processing domain with their work on the Graph Concept Explainer (GCExplainer). More recently, their work was extended toward hierarchical concepts \cite{jurssEverybodyNeedsLittle2023} and shared concept spaces for multimodal learning \cite{dominiciSHARCSSharedConcept2023a}.
In another line of work, Azzolin \etal \cite{azzolinGlobalExplainabilityGNNs2022b} introduce the Global Logic-based Graph Explainer (GLGExplainer), which uses logic explainable networks (LEN) \cite{ciravegnaLogicExplainedNetworks2023} to form predictions by combining concepts via logical formulas.

\section{Background}

\subsection{MEGAN: Multi-Explanation Graph Attention Network}
\label{sec:megan}

\begin{figure}[t!]
    \centering
    \includegraphics[width=\textwidth]{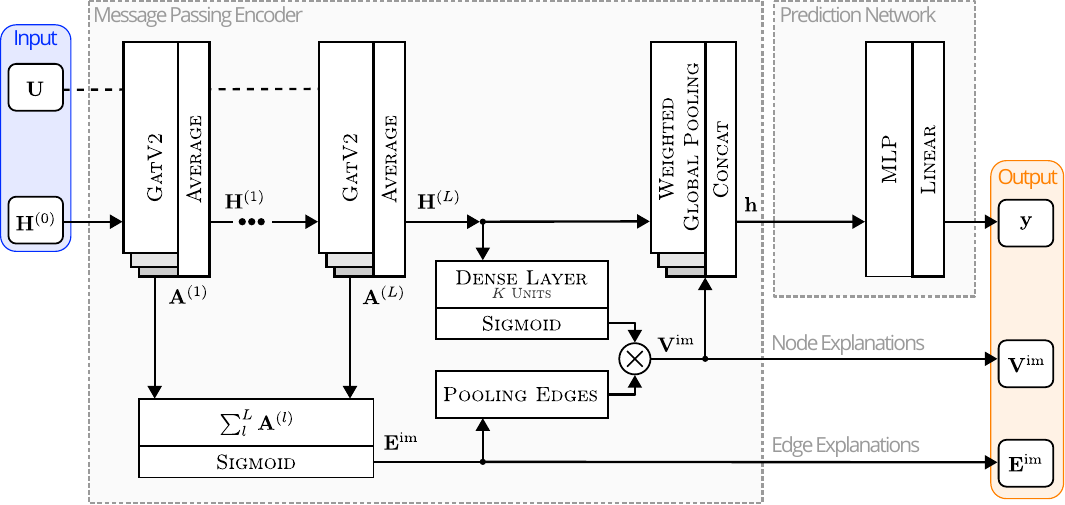}
    \caption{Simplified overview of the multi-explanation graph attention network (MEGAN) architecture. Modified illustration based on the original work of Teufel \etal \cite{teufelMEGANMultiexplanationGraph2023a}. Rounded boxes represent input/output tensor structures of the network. Square boxes represent network layers and arrows indicate layer interconnections.}
    \label{fig:architecture}
    \vspace*{-0.2cm}
\end{figure}

The multi-explanation graph attention network (\megan) was recently introduced as a self-explaining graph neural network architecture, which directly produces local node and edge attributional explanations alongside its main target predictions. The following section summarizes the main features of the \megan~architecture (see Figure~\ref{fig:architecture} for a visual overview), while a detailed description can be found in the original work by Teufel \etal \cite{teufelMEGANMultiexplanationGraph2023a}.\\
The model's attributional explanation masks are based on the model's internal attention mechanism and serve to highlight graph substructures which are particularly important for the model's prediction. Specifically, the model produces a multiple of $K$ attributional explanation masks for each element of the input graph structure, where the number of explanations is an architectural hyperparameter independent of the task specifications.\\ 
Formally, The input data structure $\inputElem = \left( \graph, \nodeAttr{0}, \edgeAttr{0} \right)$ of a \megan~model consists of a graph structure $\graph$, a tensor $\nodeAttr{0} \in \mathbb{R}^{\nodeNum \times N_0}$ of initial node features, and a tensor $\edgeAttr{0} \in \mathbb{R}^{\edgeNum \times M_0}$ of initial edge features. The graph $\graph = \left(\nodes, \edges \right)$ as a set $\nodes = \left\{0, \dots, V \right\}$ of nodes and a set $\edges \subset \nodes \times \nodes$ of node pairs. As the corresponding output $\targetElem = \left(\targetPred, \nodeImportances, \edgeImportances \right)$, the model produces the tensor of target value predictions $\targetPred \in \mathbb{R}^{\targetNum}$, the multi-channel node importance tensor $\nodeImportances \in [0, 1]^{\nodeNum \times \channelNum}$, and the multi-channel edge importance tensor $\edgeImportances \in [0, 1]^{\edgeNum \times \channelNum}$. The model can therefore be described as a function
\begin{equation}
f_{\params}: \left(\graph, \nodeAttr{0}, \edgeAttr{0} \right) \mapsto (\targetPred, \nodeImportances, \edgeImportances)
\end{equation}
with trainable parameters $\params$.\\
The message passing part of \megan's architecture is based on \gatII~ \cite{brodyHowAttentiveAre2022a} attention layers as its main attention mechanism. The first part of the network consists of $\layerNum$ attention layers, where each layer consists of $\channelNum$ parallel attention heads.
The $\layerIndex$-th attention layer---consisting of $\channelNum$ individual attention heads---outputs a node embedding representation $\nodeAttr{l}$ based on the graphs edge features $\edgeAttr{0}$ and the previous layers node features $\nodeAttr{l-1}$.
In addition to the node feature updates, each attention head also produces an edge attention tensor $\edgeAttention{l} \in \mathbb{R}^{\edgeNum \times \channelNum}$. These edge attention values are then aggregated across all layers to produce edge importance tensor $\edgeImportances \in \mathbb{R}^{\edgeNum \times \channelNum}$, representing the edge explanations. Additionally, the node importance tensor $\nodeImportances \in \mathbb{R}^{\nodeNum \times \channelNum}$ is obtained from a combination of edge importances and node embeddings of the final layer. After this message-passing part of the network, a graph pooling operation is used to create the global graph embedding $\graphEmbedding$. The graph pooling is realized as an attention-weighted sum over the final node embeddings---done for each channel independently---resulting in the channel-specific embedding
\begin{equation}
    \embedding{k} = \sum_{i}^{\nodeNum} \nodeAttr{L}_{i, :} \cdot \nodeImportances_{i, k} 
\end{equation}
which are then concatenated into the overall graph embedding $\graphEmbedding = \embedding{0} || \cdots || \embedding{K}$. This overall graph embedding is then passed through an MLP with $\targetNum$ output units, which results in the target prediction $\targetPred$.\\
%
%
To promote the model's explanation masks to behave according to their pre-determined interpretations, the explanation co-training method is introduced. In addition to the main prediction loss $\lossPred$, during training, the model is also subject to an explanation loss $\lossExpl$, as well as a sparsity regularization term $\lossSpar$. The total training loss is therefore given as 
\begin{equation}
    \mathcal{L} = \lossPred + \beta \cdot \lossExpl + \gamma \cdot \lossSpar
\end{equation}
where the weights $\beta$ and $\gamma$ are model hyperparameters.\\
The explanation co-training loss trains the network to approximate a solution to the primary target prediction problem purely based on the explanation masks $\nodeImportances$ across the $K$ explanation channels. The sparsity loss applies an L1 regularization on the explanation masks to promote sparser explanations that focus only on a subset of nodes and edges.

\subsection{Definition of Graph Concept Explanations}
\label{sec:concepts}

Concept-based explanations are a type of global explanations that aim to provide a set of generalizable rules about a model's overall behavior. In this context, a \textit{concept} is a pattern that can be present and shared among multiple concrete instances. We follow the convention introduced by Ghorbani \etal \cite{ghorbaniAutomaticConceptbasedExplanations2019} and represent a concept $\concept = \{ \fragmentElem \}$ as a set of input fragments $\fragmentElem \subset \inputElem$ of an input data structure $\inputElem$. What exactly constitutes such an input fragment, and therefore a concept itself, depends on the application domain. In the image processing domain, for example, valid input fragments would be superpixels or small image segments, and the concept be defined as the common motif these image snippets share.\\
For the graph processing domain, we define a concept 
\begin{equation}
\concept = \{ \subgraph, \;\dots \} = \left\{ ( \graph, \nodeMask, \edgeMask ), \;\dots \right\}
\end{equation}
as a set of subgraph motifs $\subgraph \subset \graph$. More specifically, we consider a \textit{soft} subgraph definition, where subgraphs are delimited by continuous node masks $\nodeMask \in [0, 1]^\nodeNum$ and edge masks $\edgeMask \in [0, 1]^\edgeNum$, rather than hard binary masks.\\
A concept explanation then seeks to associate concepts with specific influences on the model prediction to act as a set of generic rules about the model's behavior. Formally, Ghorbani \etal require a valid concept explanation to fulfill the three properties of \textit{meaningfulness}, \textit{coherency}, and \textit{importance}. This means that input fragments have to be large enough to independently make sense, the fragments of a concept have to share an identifiable commonality and the concept has to have demonstrable importance for the model's prediction.\\

\section{Methods}


We propose a framework for the generation of concept explanations for graph property prediction tasks, in which the concepts are derived as clusters in a latent space of subgraph embeddings (see Fig.\ref{fig:basic-overview}). In this context, we define a subgraph latent space such that the constituting vector representations only embed information about graph substructures rather than the full graph.

\begin{assumption}
\label{ass:subgraph}
There exists a function $f^{\dag}: \graph \mapsto \graphProjection$ that projects a graph structure $\graph \in \mathbb{G}$ to a latent vector representation $\graphProjection \in \mathbb{R}^{D}$ such that the embedding only captures information about a specific subgraph structure $\subgraph \subseteq \graph$. In other words, that $f^{\dag}(\graph^1) = f^{\dag}(\graph^2)$ when $\subgraph \subseteq \graph^1, \graph^2$ even if $\graph^1 \neq \graph_2$
\end{assumption}

Additionally, we expect such a latent space representation to reflect structural similarities between the relevant graph structures. In other words, if two graphs contain substructures that are only slightly different, then their corresponding embeddings should lie close together.

\begin{assumption}
\label{ass:similarity}
May $f^{\dag}: \graph \mapsto \graphProjection$ be a subgraph projection, then there exists at least one distance measure $d: \mathbb{R}^{D} \times \mathbb{R}^D \rightarrow \mathbb{R}$ for the embedding space that is proportional to a measure $s: \mathbb{G} \times \mathbb{G} \rightarrow \mathbb{R}$ of structural similarity between the original graphs: $d(\graphProjection^1, \graphProjection^2) \propto s(\graph^1, \graph^2)$.
\end{assumption}

Based on these assumptions, dense clusters in such a subgraph latent space represent original input graphs that are not necessarily similar overall, but which contain similarly structured subgraph motifs. Therefore one can say that the cluster members are graphs that share a common structural \textit{concept}.\\
We argue that the architecture of the \megan~graph neural network already approximately fulfills Assumption \ref{ass:subgraph}. In the model, the global graph pooling operation is realized as an attention-weighted sum, where the individual node embeddings are multiplied by their corresponding importance values (see Sec. \ref{sec:megan}). The pooled embedding vector is mostly comprised of information from high-importance nodes, while information from low-importance nodes is suppressed. Therefore, the pooled embedding $\embedding{\channelIndex}$ from the $\channelIndex$-th explanation channels mostly represents the subgraph structure defined by that channel's explanation mask.\\
In general, Assumption \ref{ass:similarity} will not be fulfilled, since the latent space will be structured according to the network's primary target predictions. In the case of a three-class classification task, for example, the latent space will be structured into three clusters that maximize the decision boundary between the classes. Therefore, additional measures are necessary to ensure Assumption~\ref{ass:similarity} to be approximately fulfilled. For this purpose, we propose the updated \meganII~version of the original model with the specific intent of facilitating the extraction of global concept explanations. 

\subsection{Extended Network Architecture}
\label{sec:projection-layers}

In the original \megan~model architecture, each channel's embedding $\embedding{\channelIndex}$ is created as the attention-weighted sum of the graph's node embeddings, then immediately concatenated and used as input to the final property prediction network. We introduce additional projection networks $g^{(\channelIndex)}: \mathbb{R}^{D} \rightarrow \mathbb{R}^P, \; \embedding{\channelIndex} \mapsto \projection{\channelIndex}$ for each channel separately, which transform the original embedding $\embedding{\channelIndex} \in \mathbb{R}^{D}$ into a projected embedding $\projection{\channelIndex} \in \mathbb{R}^{P}$. This additional projection serves two main purposes: Firstly, it ensures that each channel's embeddings can develop independently. Without the additional projection, the embeddings directly result from a summation of the node embeddings. A change of the pooled embedding $\embedding{\channelIndex}$ would be necessarily coupled to a change of the node embeddings. However, since the embeddings of all other channels also directly result from those same node embeddings, each channel's representations are strongly coupled to each other. With the additional projection, this is no longer necessary and channel embeddings can develop independently from each other. The second purpose of the projection networks is to increase the general encoding capabilities of the latent spaces. In other words, to make sure that sufficiently complex functions can be represented---which can capture greater details about subgraph similarity relationships. Specifically, by increasing the dimensionality $\projectionDim \gg \embeddingDim$ of the embeddings through the projection, we make sure that more information can be encoded in the representations.\\
Additionally, L2-normalization is applied as the activation of the projection network's output which causes all the projected embeddings $\projection{k}$ to lie on the $\projectionDim$-dimensional unit sphere. This normalization makes the cosine similarity an especially suitable distance measure for the latent space.

\begin{figure}[t!]
    \centering
    \includegraphics[width=\textwidth]{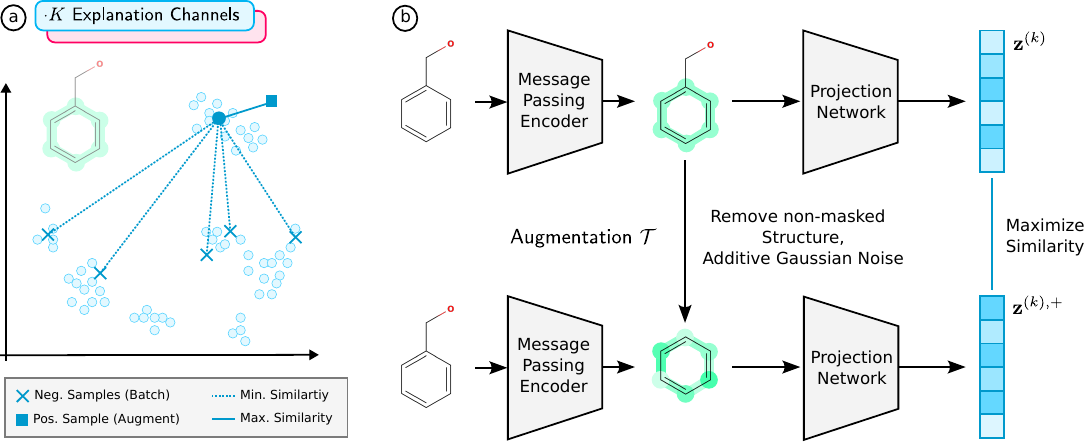}
    \caption{Visualization of the contrastive learning method. \circla~The contrastive learning is applied for each explanation channel's latent space individually by maximizing an embedding's similarity to the positive sample and minimizing similarity to negative samples. Other elements in the training batch are used for the negative samples and the positive samples are derived by data augmentation. \circlb~The augmentation is applied in the final stage of the message-passing encoder, during the global graph pooling operation. The augmented view removes all graph structures outside the explained areas and applies additive gaussian noise to the explanation masks.}
    \label{fig:contrastive}
    \vspace*{-0.2cm}
\end{figure}

\subsection{Contrastive Explanation Learning}
\label{sec:contrastive}

For the proposed concept clustering scheme to yield meaningful results, Assumption~\ref{ass:similarity} has to be approximately fulfilled, which means that the distance in the model's latent space should be proportional to the structural similarity of the corresponding graph motifs. To promote this property, we add a contrastive representation learning objective $\lossContr$ to the overall training loss
\begin{equation}
    \mathcal{\hat{L}} = \lossPred + \beta \cdot \lossExpl + \gamma \cdot \lossSpar + \mu \cdot \lossContr
\end{equation}
of the \megan~model, where $\mu$ is a hyperparameter of the model that determines the weight of this new loss term during training.\\
For the contrastive learning scheme, we follow the general SimCLR framework \cite{chenSimpleFrameworkContrastive2020} (see Fig.~\ref{fig:contrastive}), for which the basic idea is to maximize an embedding's similarity with a set of positive samples (similar input elements) while minimizing the similarity with a set of negative samples (dissimilar input elements). 
We apply the InfoNCE loss
\begin{equation}
    \mathcal{L}^{\mathrm{contr},(k)} = - \log \left( \frac{ \exp(\graphProjection^{(k)} \cdot \graphProjection^{(k),+} / \tau)}{\exp(\graphProjection^{(k)} \cdot \graphProjection^{(k),+} / \tau) + \sum_{\batchIndex}^{\batchNum} \exp(\graphProjection^{(k)} \cdot \graphProjection^{(k), -}_{\batchIndex} / \tau) } \right)
\end{equation}
separately to the projected embeddings $\graphProjection^{(k)}$ of each explanation channel. Since the projected embeddings are L2-normalized (see Sec.~\ref{sec:projection-layers}), the inner product between the embedding vectors is equivalent to their cosine similarity. 
For each element in a training batch with batch size $\batchNum$, all other elements of the batch are considered negative samples $\{ \graphProjection^{(k), -}_{\batchIndex} \}$. The positive samples $\graphProjection^{(k),+}$ are derived through data augmentations. Specifically, two different kinds of augmentations are applied to the channel's explanation mask $\nodeImportances_{:, k}$ during the forward pass of the model (see Fig~\ref{fig:contrastive}). First, we use a threshold to turn the still-continuous explanation into a binary mask. During the model forward passes, this binary mask is then applied to the node embedding during the global pooling such that all nodes outside of the mask are ignored. As a second augmentation, we apply additive Gaussian noise to the explanation mask itself. Together, these augmentations promote the embeddings to mainly represent the explained substructures, while being invariant towards small changes in the mask's specific importance values.


\subsection{Concept Clustering}

By training a \meganII~model with an additional contrastive loss term (see Section~\ref{sec:contrastive}) acting on the projected channel embeddings $\projection{k}$, Assumption~\ref{ass:subgraph} and Assumption~\ref{ass:similarity} are approximately fulfilled. Consequently, the channel projections $\projection{k}$ primarily encode information about the 
subgraph highlighted in the corresponding explanation. In addition, the distance between two latent projections approximately corresponds to the structural similarity of their motifs.
Therefore, overarching structural concepts can be identified as dense clusters within this projected latent space.\\
To find these clusters we use the HDBSCAN \cite{malzerHybridApproachHierarchical2020} clustering algorithm on the channel projections $\projection{k}$ for each of the $\channelNum$ channels independently (see Figure~\ref{fig:basic-overview}). HDBSCAN is a density-based clustering algorithm that doesn't require the number of clusters to be known beforehand. Instead, an appropriate number of clusters is dynamically discovered from the data.\\
The clustering is done for each of the model's $\channelNum$ explanation channels separately. Therefore, a concept cluster $\concept^{(k,q)}$ with index $q \in \{0, \dots, Q\}$ is also uniquely associated with the explanation channel $\channelIndex \in \{0,\dots,\channelNum\}$ from which it was extracted. The members of these concept clusters consist of various soft graph fragments $\subgraph$ (see Section~\ref{sec:concepts}).\\


Each concept cluster can additionally be associated with an average contribution to the outcome of the prediction---thereby forming an explanation of the model's behavior that associates a structural pattern (cluster members) with an effect on the predicted property (contribution).\\
For each concept cluster, this contribution is calculated as an average over the contributions of the cluster members. Each member of a cluster $\concept^{(k,q)}$ is the combination of an original graph element $\graph$ from the dataset its corresponding local explanation mask $\nodeImportances_{:,\channelIndex}$ for the $\channelIndex$-th explanation channel. Here, we define one such element's contribution as it's leave-one-out deviation $\Delta^{(k)}$ as it was originally introduced by Teufel \etal \cite{teufelMEGANMultiexplanationGraph2023a} to quantify the fidelity of \megan's explanation channels.




\begin{figure}[t!]
    \centering
    \includegraphics[width=\textwidth]{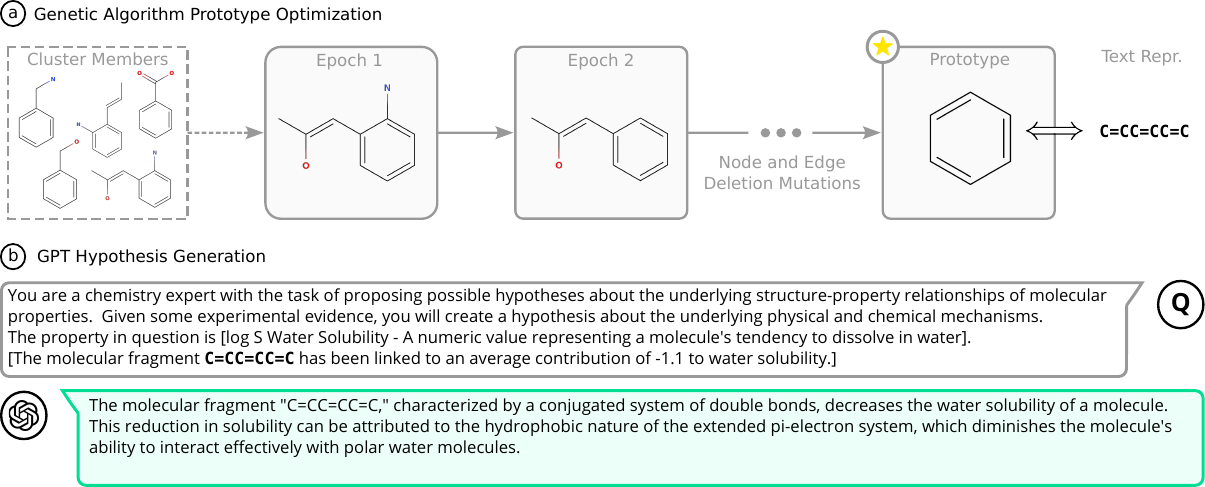}
    \caption{\circla~Illustration of the prototype optimization process. Starting from a population consisting of a concept's member graphs, node and edge deletion mutations are applied in each epoch to find a suitable prototype graph. \circlb~Example for the GPT-based hypothesis generation. The text representation of the concept prototype graph is included in a query that prompts the model to generate a causal explanation for the concept's structure-property relationship.}
    \label{fig:prototype}
    \vspace*{-0.2cm}
\end{figure}

\subsection{Prototype Optimization}

To increase the interpretability of the concept representations, we additionally generate a prototype graph $\graph^{(k, q)}_*$ for each concept cluster. In this context, we define a prototype as a graph consisting of a minimal number of nodes that is still sufficiently representative of the concept's underlying pattern. Formally, we optimize these prototype graphs 
\begin{equation}
\begin{aligned}
    \graph^{(k,q)}_* = \; & \argmin_{\graph \,\in\, \concept^{(k, q)}} \; |\graph| \\
    & \mathrm{s.t.} \quad \graphProjection^{(k)} \cdot \overline{\graphProjection}^{(k,q)} \leq \epsilon \\
    & \text{where } \graphProjection^{(k)} \text{ embedding of } \graph
\end{aligned}
\end{equation}
for each of the concept clusters separately. The main objective function of this optimization is to minimize the graph size in terms of the number of nodes. Additionally, the optimization is subject to the constraint of maintaining at least a minimum of structural similarity to the underlying pattern of the concept cluster. This constraint is realized as a minimum threshold $\epsilon$ of cosine similarity between the  

constraining the corresponding channel projection $\graphProjection^{(\channelIndex)}$ to maintain at least a minimal cosine similarity $\epsilon$ to the concept cluster centroid $\overline{\graphProjection}^{(k,q)}$. Here, the cluster centroid $\overline{\graphProjection}^{(k,q)}$ vector is computed as the average vector over all embeddings corresponding to the concept's cluster members.\\
We use a genetic algorithm (GA) with a fixed number of epochs for the optimization of the concept prototype graphs (see Figure~\ref{fig:prototype}). The initial population of the GA is chosen as the concept members' full graph structures. In each epoch, exclusively random node and edge deletion mutations are applied to all elements of the population. The GA uses a tournament selection strategy, while a small fraction of the best-performing individuals are guaranteed to be transferred to the next epoch.

\subsection{Hypothesis Generation}

If applicable, we also generate hypotheses about the causal reasons behind the extracted structure-property correlations. For this purpose, we construct a prompt containing the string representation of the prototype graph and the average contribution of the concept toward the prediction outcome. We query OpenAI's GPT-4 API to generate a causal hypothesis about why the given pattern---represented by the prototype graph---might have the associated impact on the prediction (see Figure~\ref{fig:prototype} for an example).\\
However, this hypothesis generation is only available for tasks about which the language model contains prior domain knowledge and for which meaningful text representations exist. Consequently, the method is not applicable to custom synthetic graph property prediction tasks. However, it can be used for most molecular property prediction tasks, for which the graph structure can be given the model in SMILES representation.

\section{Computational Experiments}

To illustrate the effectiveness of our proposed global explanation framework, we conduct two parts of computational experiments. In the first part, we apply our method to synthetic datasets. The labels of these datasets were created based on pre-determined structure-property relationships, where the existence of certain subgraph motifs results in corresponding graph labels. For such datasets, global explanations should be able to discover these basic structure-property relationships that were initially used to construct the dataset. In the second part, we apply our method to real-world molecular property prediction datasets. Due to the complexity of real-world tasks, there is no certainty about the true underlying structure-property relationships. Consequently, there exists no hard ground truth in terms of the generated explanations. However, for the chosen datasets, there either exist commonly known rules of thumb about certain structure-property relationships or previously published hypotheses in the scientific literature, which the generated explanations are compared with.

\subsection{Synthetic Datasets}

\begin{figure}[t!]
    \centering
    \includegraphics[width=\textwidth]{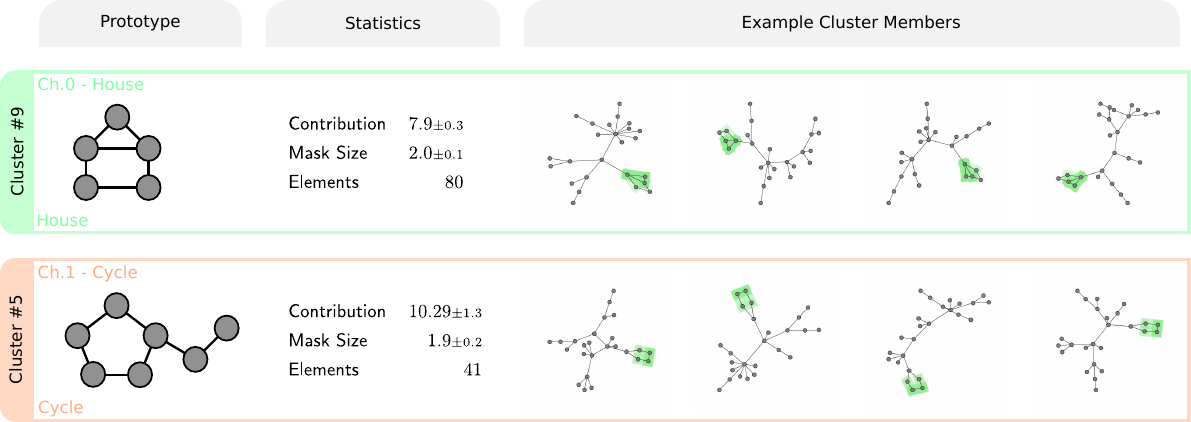}
    \caption{Selected results of the concept clustering for a \meganII~model trained on the BA2Motifs dataset. Each row represents one concept cluster for either the house class (\textcolor{SeaGreen}{green}) or the cycle class (\textcolor{Peach}{orange}). Columns from left to right show the result of the prototype optimization, aggregated statistics for the cluster members, and the 4 members closest to the cluster centroid. Prototype graphs are automatically optimized during the report generation, but have been re-drawn for the clarity of the visualization.}
    \label{fig:ba2motifs-examples}
    \vspace*{-0.2cm}
\end{figure}

\textbf{BA2Motifs - Graph Classification. } The BA2Motifs dataset was introduced by Luo \etal \cite{luoParameterizedExplainerGraph2020} and is considered a standard dataset to evaluate graph explainability \cite{kakkadSurveyExplainabilityGraph2023a}. The dataset consists of 1k randomly generated Barabási-Albert (BA) graphs. Each graph is either seeded with either a "house" motif or a five-node "cycle" motif which determines the graph's class for the underlying classification task. Therefore, the house motif and the cycle motif are considered the perfect ground truth structure-property explanations for this task.\\
In this experiment, we train a \meganII~model on the BA2Motifs dataset to generate global concept explanations and concept prototypes. The model is evaluated for its target prediction performance and explanation accuracy regarding the known ground truth explanation masks. For the performance evaluation, the model uses a 90/10 train-test split while the concept clustering is done on the full dataset.\\
%
The results firstly show that the model can perfectly solve the simple classification task with the expected test-set accuracy of 100\%. The model also scores a high explanation accuracy (node AUC=0.96, edge AUC=0.97) comparable to state-of-the-art local explainers. 
More importantly, regarding the global concept explanations (see Fig.~\ref{fig:ba2motifs-examples}), the house and the cycle pattern are correctly identified as the relevant motifs for the task. The prototype optimization is also able to recover almost the exact motifs for both cases, with only some additional nodes attached to the cycle pattern. It is also important to point out that our method returns 14 concept clusters for this dataset, indicating some redundancy in the clustering. However, all concepts exclusively represent either the house or the cycle pattern---differing only in the local neighborhood of the BA graphs where these motifs are attached.\\

\begin{figure}[t!]
    \centering
    \includegraphics[width=\textwidth]{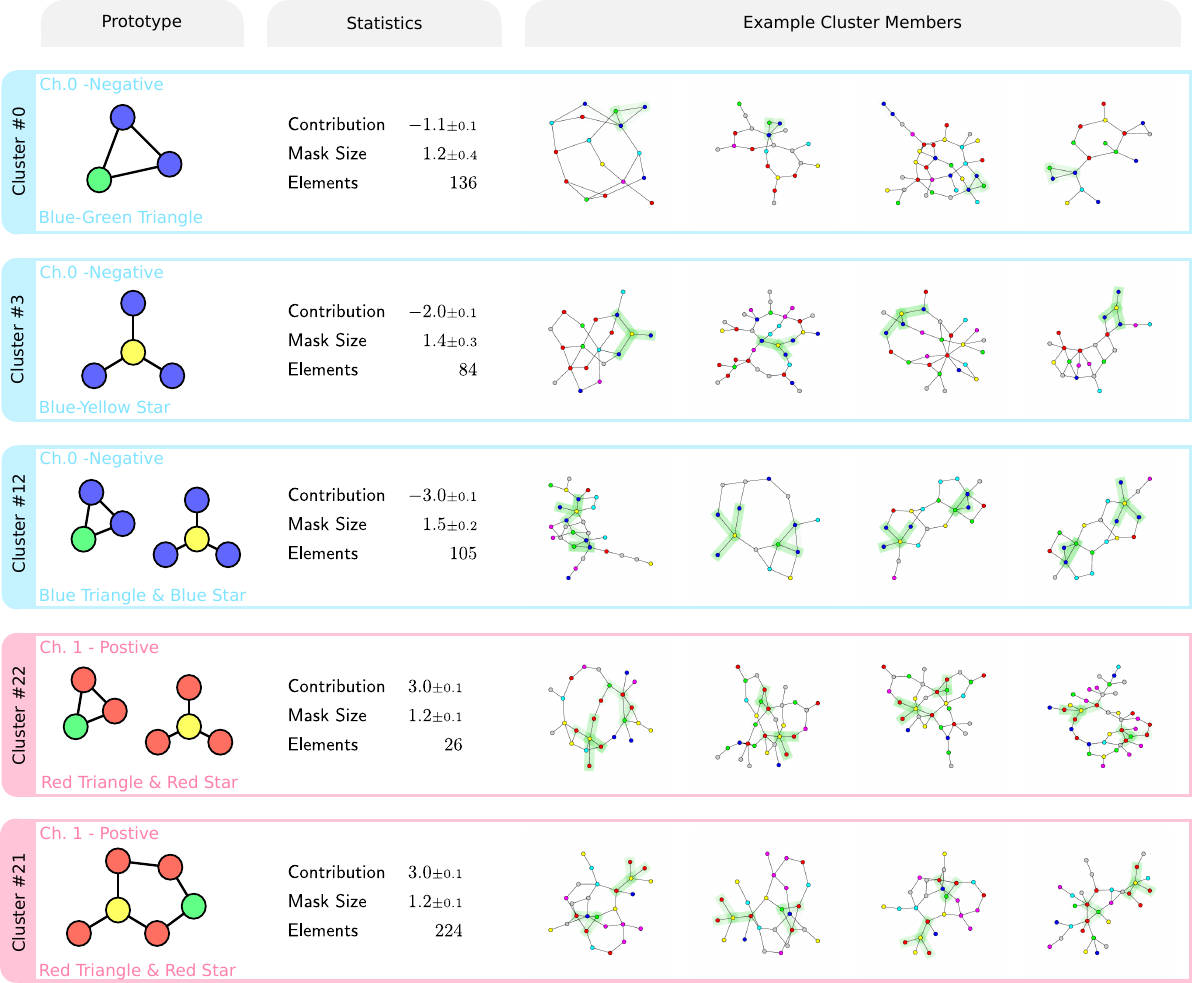}
    \caption{Selected results of the concept clustering for a \meganII~model trained on the RbMotifs dataset. Each row represents one concept cluster for either the negative (\textcolor{SkyBlue}{blue}) or positive (\textcolor{Salmon}{red}) explanation channel. Columns from left to right show the result of the prototype optimization, aggregated statistics for the cluster members, and the 4 members closest to the cluster centroid. Prototype graphs are automatically optimized during the report generation, but have been re-drawn for the clarity of the visualization.}
    \label{fig:rbmotifs-examples}
    \vspace*{-0.2cm}
\end{figure}

\begin{table}[b]
    \centering
    \caption{Evaluation results for computational experiments different model versions on the RbMotifs dataset. Experiments were repeated 5 independent times with the same train-test split. We report average results and standard deviation  for 5 independent repetitions of the dataset.}
    \label{tab:rbmotifs}
    \renewcommand{\arraystretch}{1.1}
\setlength{\tabcolsep}{6pt}

\begin{tabular}{lrrrr}

\toprule
\multicolumn{1}{c}{Model} &
\multicolumn{1}{c}{$\rsquare \uparrow$} &
\multicolumn{1}{c}{Node AUC $\uparrow$} &
\multicolumn{1}{c}{Edge AUC $\uparrow$} &
\multicolumn{1}{c}{\#Clusters $\uparrow$} \\ 

\midrule

\megan &
$0.97 {\textcolor{gray}{\mathsmaller{\pm 0.01}}}$ &
$0.99 {\textcolor{gray}{\mathsmaller{\pm 0.01}}}$ &
$0.95 {\textcolor{gray}{\mathsmaller{\pm 0.01}}}$ &
$7.8 {\textcolor{gray}{\mathsmaller{\pm 0.75}}}$ 
\\

\meganII &
$0.98 {\textcolor{gray}{\mathsmaller{\pm 0.01}}}$ &
$0.99 {\textcolor{gray}{\mathsmaller{\pm 0.01}}}$ &
$0.95 {\textcolor{gray}{\mathsmaller{\pm 0.03}}}$ &
$18.6 {\textcolor{gray}{\mathsmaller{\pm 4.13}}}$ 
\\
\bottomrule

\end{tabular}
\end{table}

\textbf{RbMotifs - Graph Regression. } The RbMotifs dataset was introduced by Teufel \etal \cite{teufelMEGANMultiexplanationGraph2023a} as an example of a regression task that is influenced by substructures of opposing influence. The dataset consists of 10k randomly generated color graphs, where the feature vector for each node represents an RGB color code. The graphs can additionally be seeded with multiple subgraph motifs---two mainly blue motifs that contribute negatively and two mainly red motifs that contribute positively to the overall graph label.\\
In contrast to other attributional explanation methods, the \megan~model is shown able to correctly capture the polarity of the opposing influences presented by the negative and positive motifs for this regression task \cite{teufelMEGANMultiexplanationGraph2023a}. However, previous work only investigates local explainability in the form of attributional masks for each individual prediction. While this already reveals some information about the underlying task, purely local explanations still require a manual review of many individual examples to be able to spot the underlying motifs that govern the task as a whole.\\
In this experiment, we train a \meganII~model on the RbMotifs dataset to generate global concept explanations and concept prototypes. Additionally, we train a model with the original unmodified \megan~architecture to compare against. Models are evaluated on their target prediction performance and the explanation accuracy with respect to the ground truth explanation masks. For the performance evaluation, both models use the same 90/10 train-test split while the concept clustering is done on the full dataset.\\
We most importantly find that the \meganII~model shows comparable prediction performance ($\rsquare \approx 0.98$), and explanation accuracy (node AUC $\approx 0.99$, edge AUC $\approx 0.95$) as the original \megan~model (see Tab.~\ref{tab:rbmotifs}). This indicates that our proposed modifications do not interfere with the model's primary predictive and explanatory function. Regarding the concept clustering, \meganII~results in significantly more clusters (18 vs 7) with the same clustering parameters.\\
Regarding the concept clustering report that is generated for the \meganII~model (see Fig.~\ref{fig:rbmotifs-examples}), we find that all of the structure-property relationships from which the dataset was initially constructed are reflected in the concept explanations, almost perfectly matching each motif's ground truth contribution to the overall graph target value. For example, there exist concept clusters that correctly identify the blue-green triangle (true contribution -1, predicted -1.1) and the blue-yellow star (true contribution -2, predicted -2) pattern as the relevant motifs related to this task. Furthermore, a different cluster can be found which features both patterns (e.g. Fig.~\ref{fig:rbmotifs-examples}, Cluster\#12) with a correctly predicted contribution of -3---therefore correctly representing the additive nature of the task.\\
Overall, the method shows coverage of the ground truth explanatory motifs, which is \textit{complete} and \textit{comprehensive}, meaning that the report contains all of the relevant motifs and no additional irrelevant ones. However, we find that the method tends to \textit{reduncancy} in the sense that the same motif can be the subject of multiple clusters. For example, there exist multiple clusters related to the combined red-triangle and red-star pattern, as illustrated in Figure~\ref{fig:rbmotifs-examples} for Cluster\#21 and Cluster\#22. This example also shows that duplicate clusters can have slightly different prototypes. While Cluster\#21's prototype shows the correct motifs, Cluster\#21's prototype shows a slightly merged version of the two motifs.


\subsection{Real-World Datasets}

\begin{figure}[t!]
    \centering
    \includegraphics[width=\textwidth]{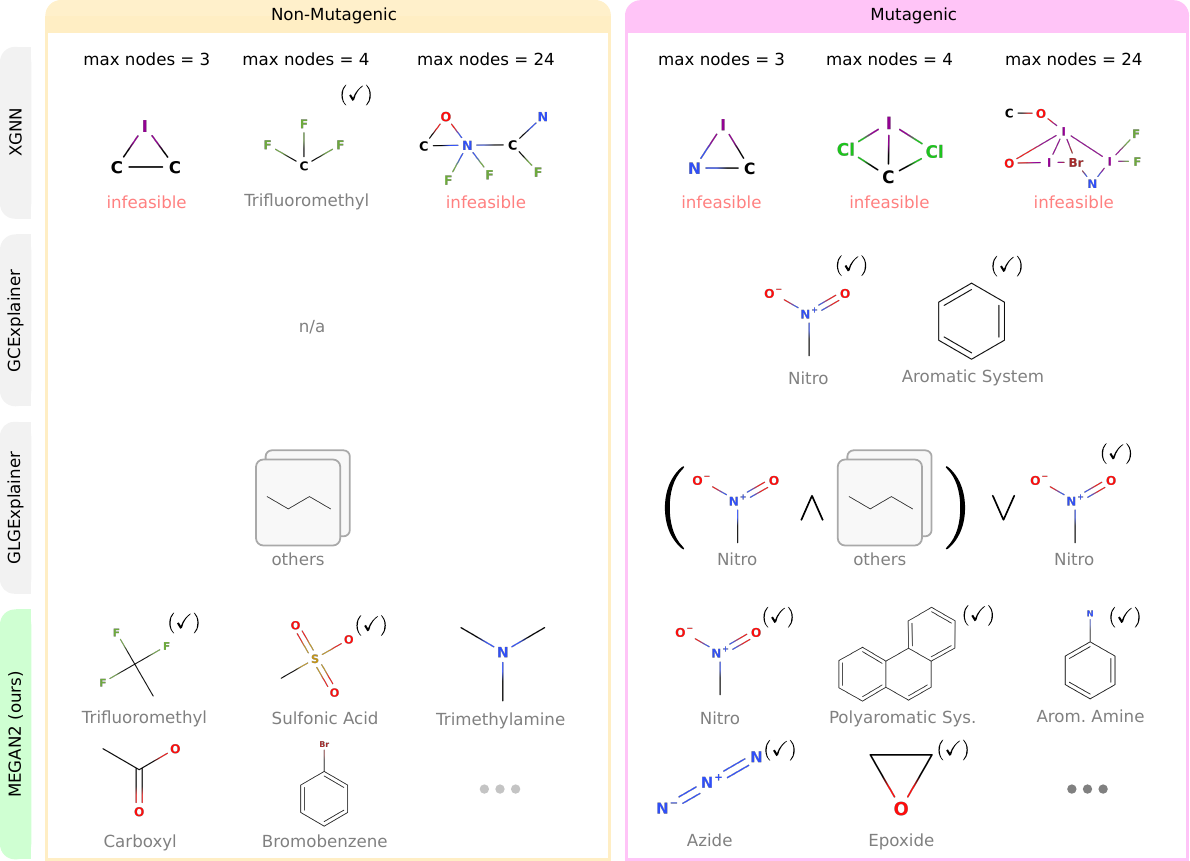}
    \caption{Comparison of global graph explainability methods for the Mutagenicity dataset. The two columns represent the two possible classes non-mutagenic (\textcolor{Apricot}{yellow}) and mutagenic (\textcolor{CarnationPink}{purple}). Each row shows the global explanations of a different method from the literature, while the last row shows explanations obtained by our proposed concept clustering method based on the \meganII~model. Explanations for XGNN and GLGExplainer are taken from Azzolin \etal \cite{azzolinGlobalExplainabilityGNNs2022b} and explanations for GCExplainer are taken from Magister \etal \cite{magisterGCExplainerHumanintheLoopConceptbased2021}. \\ \hspace*{8pt} $ ^{(\checkmark)}$ Symbol indicates consistency with hypotheses from chemistry literature \cite{kaziusDerivationValidationToxicophores2005}}.
    \label{fig:mutagenicity-comparison}
    \vspace*{-0.2cm}
\end{figure}

\textbf{Mutagenicity - Graph Classification. } The mutagenicity dataset \cite{hansenBenchmarkDataSet2009a,martinezAmesMutagenicityDataset2022} consists of roughly 6500 molecular graphs annotated with a binary classification label of being either mutagenic or non-mutagenic. These labels have been experimentally determined by test results of Ames mutagenicity assays \cite{amesImprovedBacterialTest1973,mortelmansAmesSalmonellaMicrosome2000}, which measure a substance's capability to induce mutations to the DNA of bacteria.\\
In this experiment, we train a \meganII~model on the Mutagenicity dataset to generate global concept explanations and concept prototypes. Training and evaluation of the model are done on a random 90/10 train-test split of the dataset, while concepts are extracted on the full dataset. We also compare the extracted concepts with explanations from previously published work on global graph explainability---namely XGNN \cite{yuanXGNNModelLevelExplanations2020}, GCExplainer \cite{magisterEncodingConceptsGraph2022a}, and GLGExplainer \cite{azzolinGlobalExplainabilityGNNs2022b}. XGNN is a generative explanation method, which uses a reinforcement learning approach to generate graphs that maximize classification probability. GCExplainer and GLGExplainer are concept-based explainability methods. For GLGExplainer specifically, individual motifs can additionally be joined to express explanations as logic formulas.\\
When comparing the different explainability methods for the mutagenicity prediction (see Fig.~\ref{fig:mutagenicity-comparison}), the previously published global graph explainers are subject to some limitations. Firstly we find XGNN's graph generation procedure does not take into account the chemical constraints on the graph structure, such as the maximum number of possible bonds for certain atom types. This generally leads to chemically infeasible graph structures---limiting the overall usability of the generated explanations. Since GCExplainer and GLGExplainer are concept-based explainability methods, they are not subject to the same problem. Both methods successfully recover the nitro (\ch{NO2}) group as the most relevant motif. In addition, GCExplainer also indicates aromatic carbon ring systems as possible explanations for the mutagenic class. Although these are among the most commonly considered explanations in the xAI literature \cite{yingGNNExplainerGeneratingExplanations2019b}, previous work from chemistry literature suggests that many more structural explanations for mutagenicity exist \cite{kaziusDerivationValidationToxicophores2005}. Kazius \etal \cite{kaziusDerivationValidationToxicophores2005} propose 39 structural motifs linked to mutagenic behavior, including but not limited to the often cited nitro (\ch{NO2}) group. Interestingly, the authors also mention the possibility of detoxifying structures, which can inhibit mutagenic behavior by reducing DNA interactions of otherwise mutagenic substances. We argue that these detoxifying substructures can be interpreted as explanatory motifs for the non-mutagenic class. This perspective of detoxifying structural influences was largely neglected by previous methods.\\
In contrast to existing methods, we find that our proposed concept extraction yields a substantially more diverse set of explanatory concepts for both the mutagenic and the non-mutagenic classes. For the non-mutagenic class, our method highlights, among other things, the trifluoromethyl (\ch{CF3}) and sulfonic acid (\ch{SO3H}) groups, which were specifically mentioned by Kazius \etal as possible detoxifying influences. For the mutagenic class, our method also identifies the nitro group and the polyaromatic systems as strong influences for mutagenic behavior. In addition, our method also identifies aromatic amine groups (\ch{NH2}), azide groups (\ch{N3}) and epoxide (\ch{C2O}) groups as mutagenic influences---all of which are mentioned by Kazius \etal as well. Besides the groups indicated in the literature, our method yields several additional concepts. However, we also identified some faults in the explanations. For example, while haloalkenes are correctly identified as a relevant concept for the mutagenic class, there also exists a high-fidelity haloalkene concept for the non-mutagenic class as well. We suspect that such cases of double attribution may be an artifact of the approximate assumption of linearly contributing evidence for tasks in which this is not entirely the case.\\




\begin{figure}[t!]
    \centering
    \includegraphics[width=\textwidth]{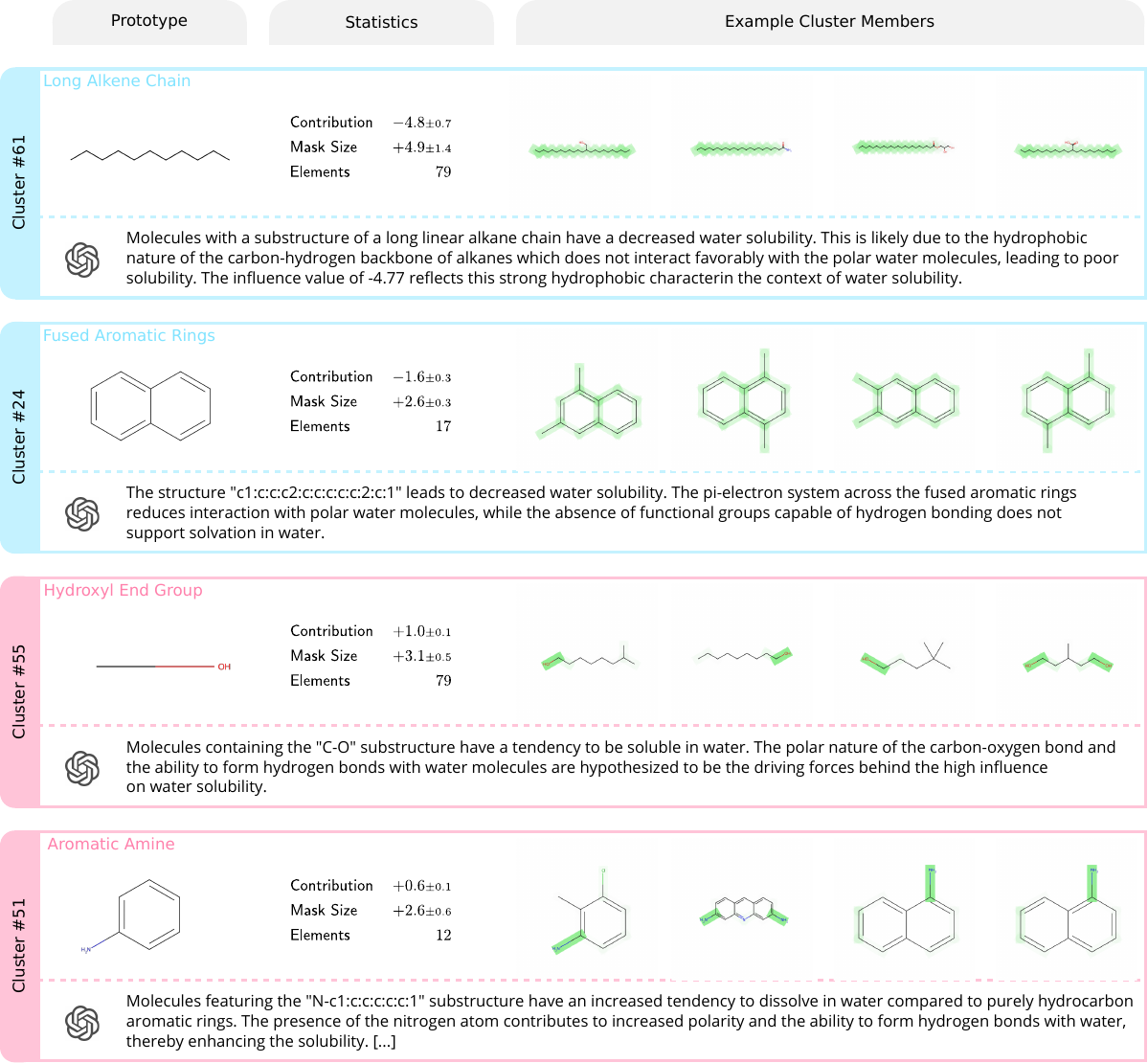}
    \caption{Selected results of the concept clustering for a \meganII~model trained on the AqSolDB dataset. Each row represents one extracted concept cluster for either the negative (\textcolor{SkyBlue}{blue}) or positive (\textcolor{Salmon}{red}) explanation channel. Columns from left to right show the result of the prototype graph optimization, aggregated statistics for the cluster members, and the 4 members closest to the cluster centroid. In addition, the GPT-generated hypothesis is displayed underneath each concept. The prototype graphs in the first column have been automatically generated.}
    \label{fig:aqsoldb-examples}
    \vspace*{-0.2cm}
\end{figure}

\begin{figure}[t!]
    \centering
    \includegraphics[width=\textwidth]{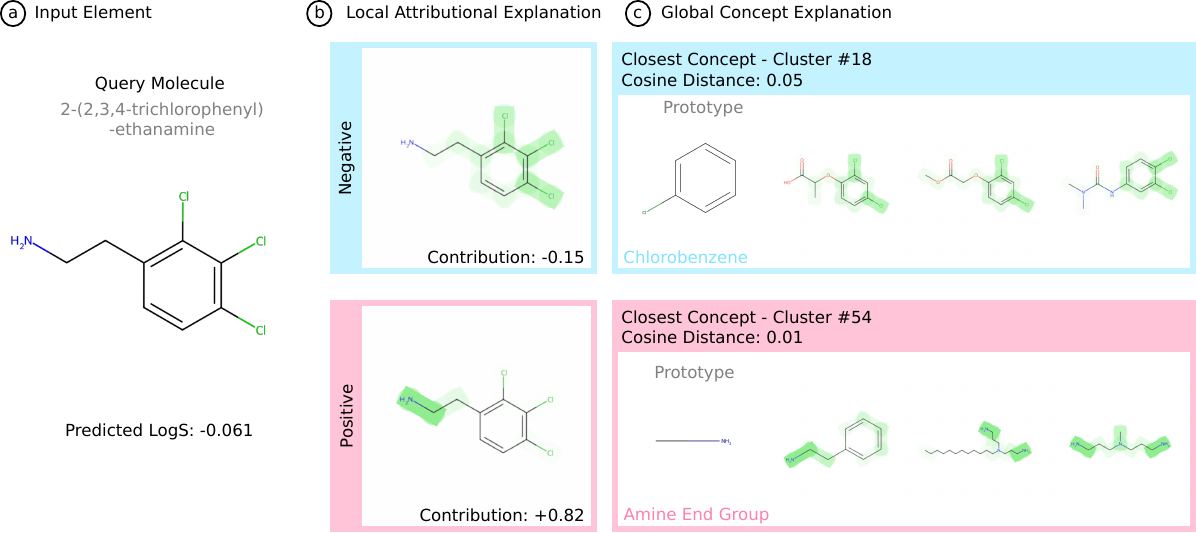}
    \caption{Visualization of how the global concept explanations can be used to improve local explainability as well. \circla~The model outputs a prediction for an individual query element. \circlb~In addition to the target value prediction itself, the model also outputs the local explanation masks for the different channels. \circlc~Each local explanation mask can be associated with a concept by finding the concept whose cluster centroid is the closest to the corresponding explanation embedding.}
    \label{fig:aqsoldb-query}
    \vspace*{-0.2cm}
\end{figure}

\textbf{AqSolDB - Graph Regression. } AqSolDB is a dataset that was introduced by Sorkun \etal \cite{sorkunAqSolDBCuratedReference2019} which consists of roughly 10k molecular graph structures annotated with experimentally determined logS values for water solubility. The values approximately lie in the range between -14 and +3, where higher values indicate a higher solubility in water.\\
For the prediction of water solubility there generally exists no exact ground truth explanations. However, since water solubility is a relatively well-studied phenomenon overall, there exist several rules of thumb about what kinds of substructures influence this property. 
In a simplified manner, water solubility is largely influenced by the polarity of a molecule and its capability to form hydrogen bonds. This is because water molecules themselves are strong dipoles and largely bind via hydrogen bonds. As a consequence, large non-polar structures such as long carbon chains and carbon rings are generally considered to reduce water solubility, while polar functional groups such as those including oxygen and nitrogen are considered to increase water solubility.\\
In this experiment, we train a \meganII~model on the AqSolDB dataset to predict the continuous logS values. The model is trained and evaluated on a 90/10 train-test split, while the concept clustering is done on the full dataset.\\
We find that our concept clustering results for the solubility prediction (see Fig~\ref{fig:aqsoldb-examples}) are generally in agreement with the previously mentioned rules of thumb. For the negative channel, the extracted concept clusters primarily represent different variants of aromatic structures, but several clusters indicate carbon chains in various sizes as another negative influence as well. For the positive channel, the extracted concepts span a variety of different functional groups mainly based on oxygen and nitrogen atoms. For example, we find the hydroxyl (\ch{OH}) and amine (\ch{NH2}) groups as two primary motifs for positive contributions to water solubility. One interesting observation regarding these results is that concepts that are described by the clusters seem to be neighborhood-dependent. An illustrative example for this is that two separate clusters describe the amine (\ch{NH2}) group. In one cluster the amines are end pieces of carbon chains (see Fig.~\ref{fig:aqsoldb-examples}, Cluster\#54) while in the other cluster, the amines are connected to aromatic systems (see Fig.~\ref{fig:aqsoldb-examples}, Cluster\#51). This behavior contradicts Assumption~\ref{ass:similarity} and most likely stems from the fact that the thresholding augmentation for the contrastive learning (see Sec.~\ref{sec:contrastive}) is applied only after neighborhood information was already spread by the message-passing. However, in the context of molecular property prediction, this neighborhood dependency is a desirable property. While the basic explanation itself is the amine group in this case, the exact way in which such a functional group affects the overall property of the molecule is also dependent on the local environment where that functional group is attached.\\
For the water solubility experiments, we furthermore use GPT-4 to generate concept hypotheses (see Fig.~\ref{fig:aqsoldb-examples}). For each concept cluster, the model is queried with a custom prompt that provides information about the identified concept and prompts the language model to form a hypothesis for the causal reasoning about why the explained substructure may be associated with a particular positive or negative influence. In some cases, these explanations seemingly reflect the known rules of thumb quite well such as for the non-polar alkene chain (Fig.~\ref{fig:aqsoldb-examples}, Concept\#61) and the polar hydroxyl group (Fig.~\ref{fig:aqsoldb-examples}, Concept\#55). However, we also observe cases where the generated hypotheses are incorrect. One possible reason for this is that the language model does not correctly identify the given motif. An example of this is Cluster\#52, whose prototype clearly shows an aromatic amine group (nitrogen attached outside of carbon ring) but is incorrectly identified as a pyridine (nitrogen part of carbon ring) group. Another possible source of error is cases where either the concept or the prototype themselves are incorrect. In these cases, the language model still tries to come up with a hypothesis about the behavior, hallucinating sentences that may sound like valid explanations but convey incomplete or incorrect content \cite{tianFinetuningLanguageModels2023}.\\

Besides shedding light on the model's overall decision-making behavior, the extracted concept explanations can also be used to improve local explainability as well. When a \megan~model is queried with a specific input graph, the model outputs the primary target prediction and the local explanations in the form of node and edge importance masks for all the explanation channels. Additionally, each explanation can be associated with one of the extracted global concepts by finding the concept cluster centroid closest to each explanation channel's corresponding embedding vector. For the water solubility prediction of the exemplary 2-(2,3,4-trichlorophenyl)ethanamine molecule (see Fig.~\ref{fig:aqsoldb-query}) the local explanations already indicate the ring structure of the graph as a negative influence and the nitrogen group as a positive influence. However, only the association with the corresponding concept explanations provides the validation that the highlighted chlorophenyl and amine functional groups are recurring patterns that are generally associated with negative/positive influences across the whole dataset. This provides further insight and helps to better understand the underlying structure-property relations

\section{Limitations}

Despite the encouraging results obtained by the proposed method, it is important to point out its limitations as well. Most importantly, since the method is based on the \megan~model, it inherits all prior limitations thereof. \megan's explanation co-training procedure approximately assumes global linearity of explanations---meaning that for an explanation to be recognized as such, a non-zero statistical correlation with the explained property has to exist. For instance, for a structure to be recognized as a "negative" explanation, it has to appear more frequently in elements with relatively low target values. While this is a reasonable assumption for many real-world tasks, it is not necessarily applicable to all kinds of graph property prediction tasks.\\
Other limitations are related to the HDBSCAN clustering algorithm that is used to find the concept clusters. Since the method is based on the density of the latent space, a certain minimum dataset size is required to estimate this density, rendering the method less suitable for small datasets. While the method is seemingly able to capture more fine-grained structural details in the identified concept clusters, it also tends to produce redundant clusters, where we sometimes observe multiple clusters that seemingly represent the same motif.\\
Another important limitation relates to the language model-based automatic generation of causal hypotheses. While these hypotheses seem to work reasonably well for some tasks, it is fundamentally restricted to tasks about which the language model has prior knowledge. For example, a model may provide useful information about relatively well-studied tasks, such as water solubility, which were likely well represented in its training corpus. However, the models will provide decreasingly useful information about niche topics or custom prediction tasks. Additionally, the accumulation of errors from prior processing steps may lead the language model to hallucinate a correct-sounding yet factually incorrect explanation. This presents a major danger, especially to non-expert users who are unlikely to recognize the errors.


\section{Conclusion}

In this work, we introduce a framework for the generation of global concept-based explanations for graph regression and classification tasks. In our framework, concepts are identified by clustering in a model's latent space of subgraph embeddings. We show that the recently proposed multi-explanation graph attention network (\megan) approximately creates such a substructure latent space. We introduce an updated \meganII~model version which includes modifications to the original architecture and training procedure to improve this latent space clustering capability. \\
We conduct computational experiments on multiple synthetic and real-world graph property prediction datasets to validate the effectiveness of the proposed explanation procedure. For the synthetic tasks, our method can correctly reproduce the basic structure-property relationships from which the datasets were originally constructed. For the real-world datasets, we find that our method rediscovers multiple known rules of thumb about molecular property regression and classification tasks. Specifically for mutagenicity prediction, our method produces substantially more fine-grained explanations than existing global graph explainability methods. Existing methods primarily focus on the nitro group as an explanation for mutagenic behavior. Contrary to that, our method recovers multiple additional structural motifs for both classes which are consistent with previous work in chemistry literature.\\
Overall, we argue that global explainability is an important step in extracting human-interpretable knowledge from complex graph property prediction models. The proposed method shows promising potential to reconstruct a task's underlying structure-property relationships. We also find that recent advances in large language models offer intriguing possibilities to communicate the extracted knowledge in natural language---which could especially benefit non-expert users and educational settings. However, in this direction, more work is needed to ensure the integrity of the generated text and prevent the spread of misinformation. Based on the promising results, future work will need to investigate the method's applicability to more complex real-world graph property prediction tasks, as well as a user study to evaluate the effectiveness of the generated explanations.\\

\textbf{Reproducibility Statement. }
All code is publicly available on GitHub: \url{https://github.com/aimat-lab/megan_global_explanations}. The code is implemented in the Python 3.10 programming language and builds on the Pytorch Geometric \cite{feyFastGraphRepresentation2019} graph neural network library. Besides the experiment modules, the repository contains the automatically generated full concept clustering reports for all of the experiments that were referenced in this publication.\\


\textbf{Acknowledgements. } This work was supported by funding from the pilot program Core-Informatics of the Helmholtz Association (HGF).\\


\textbf{Conflict of Interest. } The authors have no competing interests to declare that are relevant to the content of this article.

%

\bibliographystyle{splncs04}
\bibliography{main}

\end{document}